# Algorithms for certain classes of Tamil Spelling correction

Authors: Muthiah Annamalai*, T. Shrinivasan
Corresponding author: ezhillang@gmail.com

## 1. Introduction

Tamil language has an agglutinative, diglossic, alpha-syllabary structure which provides a significant combinatorial explosion of morphological forms all of which are effectively used in Tamil prose, poetry from antiquity to the modern age in an unbroken chain of continuity.

However, for the language understanding, spelling correction purposes some of these present challenges as out-of-dictionary words. In this paper the authors propose algorithmic techniques to handle specific problems of conjoined-words (out-of-dictionary) e.g. தென்றல்காற்று = தென்றல் + காற்று when parts are alone present in word-list in efficient way. Morphological structure of Tamil makes it necessary to depend on synthesis-analysis approach and dictionary lists will never be sufficient to truly capture the language.

### 1.1 State of the art

Many popular spell-checking applications have advanced state of the art in Tamil spelling correction but their methods are sometimes opaque. The work of Rajaraman(Vaani) [14] and Dr. Vasu Renganathan [13] remain popular in the interwebs, while Google has used hunspell/aspell variants with affix dictionary of Thamizha collaborative [15] to create spelling support of Tamil text in Google Docs product - without contributing back much. Whereas, the input method editor for Apple iOS and such from Murasu Anjal team is proprietary [16]. Our own work of solthiruthi is nascent and performs somewhat slowly due to overzealous searches, and not ready for production; however, that does not mean inferiority in class of corrections it is capable of - some of which is shared in this paper.

### 1.2 Overview

Authors propose some algorithms to identify and correct conjoined words; algorithms forMayangoli letter transposition/substitution error correction; algorithms are proposed for correcting typographical errors originating from keyboard layouts. Authors also propose using machine-learning / deep-neural network based techniques to (a) identify a Tamil word sequence as valid word or mis- spelled word. Tamil misspelled letters can be collected using crowd-sourced format and used as training data for a Tamil based Deep Neural Network (DNN) for the purpose of classifying correctly spelled and misspelled words. Further, Tamil verb declension problem can also perhaps be resolved using the big-data AI/ML approaches.

We also present algorithmic techniques to optimize a spell-checker implementation, to detect the presence of Tamil unicode character in unicode-point quickly, and also algorithm to detect foreign- language words in Tami and substitute with equivalent (when available, or acceptable Tamil form). While some of these techniques are common, this work shows the algorithms in Tamil language context, and their novelties are elucidated in this work.

## 2 Edit Distance Search Algorithm

Basic spell checker algorithm for Tamil is laid out in [1] by team of Prof. Geetha, Dhanabalan and co-workers which includes verb declension, affix removal, morpheme extraction and then applying corrections to root word and then synthesizing it.

### 2.1 Norving Algorithm

A simpler algorithm for correcting Tamil words is to directly apply the Edit distance algorithm - popularly referred to as the Norving-algorithm [2,4]. This algorithm essentially computes for each letter in a word the possibility that the letter at the position in word could be,
   1. Deleted,
   2. Substituted - with alternate letter
   3. Inserted - with alternate letter

and enumerates through all the alternate forms of input word at edit distance of 2

Clearly the presence of 247 unique Tamil letters (323 including Grantha letters) including the alpha-syllabary forms explodes the search space for a N-letter word can go as, $(247)^N$ or $(323)^N$ as you may choose to implement the algorithm for an edit-distance of upto N letters in word. Clearly not all letters in a word will be mis-spelled.

Generally the algorithm is limited to searching 2 or 3 edit distance. For an N letter word, we have then $_NC_2$ x 323 x 323 options or $_NC_3$ x 323 x 323 options respectively for the chosen edit-distance. Judiciously implementing this algorithm is key to having a data-driven spell-checker perhaps by tree-pruning techniques [11].

The implementation in Open-Tamil solthiruthi module is the following which generates suggestions which are filtered against a dictionary (typically represented as Trie data structure [8]. This edit-distance search occurs in spelling correction in Mayangoli correction, and Typographical error correction as well as described in the sections below.

```python
def norvig_suggestor(word:list,alphabets=tamil_letters,nedits=1,limit=float("inf")):
    # recursive method for edit distance > 1
    if nedits > 1:
        result = []
        for nAlternate in norvig_suggestor(wordL,alphabets,nedits-1,limit-len(result)):
            if len(result) > limit:
```

```
            break
        result.extend( norvig_suggestor(nAlternate,alphabets,1,limit-len(result)) )
    return set(result)

ta_splits     = [ [u"".join(wordL[:idx-1]),u"".join(wordL[idx:])] for idx in range(len(wordL) + 1)]
#pprint( ta_splits )
ta_deletes    = [a + b[1:] for a, b in ta_splits if b]
ta_transposes = [a + b[1] + b[0] + b[2:] for a, b in ta_splits if len(b)>1]
ta_replaces   = [a + c + b[1:] for a, b in ta_splits for c in alphabets ]
ta_replaces2  = [ c + b for a, b in ta_splits for c in alphabets ]
ta_inserts    = [a + c + b     for a, b in ta_splits for c in alphabets]
# TODO: add a normalizing pass word words in vowel+consonant forms to eliminate dangling ligatures
return set(ta_deletes + ta_transposes + ta_replaces + ta_replaces2 + ta_inserts )
```

## 2.2 Driver Algorithm for Spell Checker

சொல்திருத்தியில் கணினி நிரல் செய்யவேண்டியது இதுவே:

1. உள்ளூடு கொடுக்கப்பட்ட சொல் சரியானதா, அல்லது தவறானதா ?
2. தவறான சொல் என்ற பட்சத்தில் அதன் மாற்றங்கள் என்னென்ன ?

As per the above algorithm, we correct only words not in dictionary. This is non-word error correcting algorithm - i.e. it corrects only words not in dictionary. This algorithm won't correct word-sense disambiguation type errors, homonyms etc. - i.e. the in-dictionary word error which occurs out of place in text. Algorithm will iterate through the words in text and generate alternates using best effort for the wrong word and send them to user (replacing per user choice).

## 3. Mayangoli letter transposition/substitution error correction

தமிழில் உள்ள மயங்கொலி எழுத்துகள் நான்கு வரிசையில் அமைக்கலாம் [9],

- ல, ழ, ள வரிசை.
- ர, ற வரிசை.
- ந, ன, ண வரிசை.
- ங, ஞ, வரிசை.

சொல்திருத்தியில் கணினி நிரல் செய்யவேண்டியது இதுவே:
1. உள்ளூடு கொடுக்கப்பட்ட சொல் சரியானதா, அல்லது தவறானதா ?
2. தவறான சொல் என்ற பட்சத்தில் அதன் மாற்றங்கள் என்னென்ன ?

முதல் படியை எளிதாக ஒரு கையகராதியை கொண்டு செயல்படுத்தலாம். இதனை [ஓபன்-தமிழ் (open-tamil) solthiruthi](#) தொகுப்பில் Tamil VU மின் அகராதியை கொண்டு செயல்படுத்தியுள்ளோம். சரியான சொற்கள், அதாவது வேர் எடுத்த, புணர்ச்சி மற்றும் சாந்தி பிரிக்கப்பட்ட சொற்கள் அனைத்தும் சராசரி மின்அகராதியில் காணலாம். இதுவே எளிதான படி.

இரண்டாவது படிதான் ஒரு சொல்திருத்தியின் சிறப்பிற்கும், தரத்திற்கும், முக்கியமானது; இந்த பதிவில் எப்படி மயங்கொலி எழுத்து பிழைகளை திருத்தலாம் என்று சில எண்ணங்களை சமர்ப்பிக்கிறேன். உதாரணம் உரையின் சொல் "**பளம்**" என்பது பிழை என்று கண்டறியப்பட்டது. இது **பள்ளம்**, அல்லது **பழம்** என்று இரு மாற்றங்களை எழுத்தாளர் நினைத்தாலும் இதனை பிழையாக உள்ளீடு செய்துள்ளார். இங்கு **ள-ல-ழ** மயக்கம் காணப்படுகிறது. இதனை கணினி "பலம்", "பழம்" என்றும் மாற்றுகளை உருவாக்கி இதில் அகராதியில் உள்ளவற்றை மட்டுமே வடிகட்டி எழுத்தாளருக்கு பரிந்துரை செய்யவேண்டும். இதனை கொண்டு அணைத்து மயங்கொலி பிழைகளை திருத்தும் ஒரு தன்மை கொண்ட சொல்திருத்தியை உருவாக்கலாம். உதாரணம்,

வளர்ச்சி நிலையில் உள்ள, தற்போது மென்பொருள் வடிவமைப்பில் உள்ள சொல்திருத்தி ஓபன்-தமிழ் தொகுப்பில் காணலாம்: [எச்சரிக்கை: இது இன்னும் பொது பயன்பாட்டிற்கு பொருத்தமானதல்ல]

```
:~/devel/open-tamil$ ./spell.sh -i
>> பளம்
சொல் "பளம்" மாற்றங்கள்
(0) பம், (1) பளகு, (2) உளம், (3) பள், (4) அளம், (5) ஆளம், (6) பழம்
```

இதன் செயல்முறை கீழ்கண்டவாரு பைத்தன் மொழியில் அமையும் (இடம் சுருக்கத்தின் காரணமாக constructor மற்றும் சில comment/குறிப்புகள் தரப்படவில்லை:

```python
class Mayangoli:
    varisai = [[ u"ல்", u"ழ்",u"ள்"],[u"ர்", u"ற்"],[u"ந்",u"ன்",u"ண்"],[u"ங்",u"ஞ்"]]#வரிசை.
    @staticmethod
    def run(word,letters):
        obj = Mayangoli(word,letters)
        obj.find_letter_positions()
        if len(obj.matches_and_positions) == 0:
            return []
        obj.find_correspondents()
        obj.generate_word_alternates()
        return obj.alternates

    def find_letter_positions(self):
        for idx,letter in enumerate(self.letters):
            p = tamil.utf8.splitMeiUyir(letter)
            if len(p) == 1:
                continue
            mei,uyir=p
            for r in range(0,len(Mayangoli.varisai)):
                for c in range(0,len(Mayangoli.varisai[r])):
                    if mei == Mayangoli.varisai[r][c]:
                        self.matches_and_positions.append((idx,r,c))
        return len(self.matches_and_positions) > 0
    def find_correspondents(self):
        for pos,r,c in self.matches_and_positions:
```

```python
            src_letter = self.letters[pos]
            _,src_uyir = tamil.utf8.splitMeiUyir(src_letter)
            alt_letters = list()
            for alternate_mei in Mayangoli.varisai[r]:
                alt_letters.append( tamil.utf8.joinMeiUyir(alternate_mei,src_uyir) )
            self.pos_classes.append(alt_letters)
        return True
    def _generate_combinations(self):
        return itertools.product(*self.pos_classes)
    def generate_word_alternates(self):
         for position_sub in self._generate_combinations():
            alt_letters = copy.copy(self.letters)
            if _DEBUG: pprint.pprint(position_sub)
            idx =0
            for pos,r,c in self.matches_and_positions:
                alt_letters[pos] = position_sub[idx]
                idx += 1
            word_alt = u''.join(alt_letters)
            self.alternates.append(word_alt)
         return True
```

# 4. Algorithm for conjoined word recognition

[தென்றல்காற்று = தென்றல் + காற்று ] இரு சொற்களும் சொல் அகராதியில் இருந்தாலுன் கூட இணைந்த சொல் அகராதியில் இருக்காது. இதனை தற்காலிக சொல்திருத்திகள் பிழை என்று சொல்லும். ஆனால் இந்த செயல்முறையினால் நாம் இவற்றை பிரித்துப் பார்த்து சரியான சொல் என்று கண்டறியலாம்.

இந்த செயல்முறை ஓப்பன் தமிழ் தொகுப்பில் இவ்வாறு:

```python
class OttruSplit:
    """ யாரிகழ்ந்து = [ய்  + ஆரிகழ்ந்து], [யார், இகழ்ந்து] ,[யாரிக், அழ்ந்து], [யாரிகழ்ந்த்,உ]"""
    def run(self,lexicon):
        self.generate_splits()
        return self.filter(lexicon)
    def generate_splits(self):
        """
            யாரிகழ்ந்து =
                [['ய்', 'ஆரிகழ்ந்து'],
                ['யார்', 'இகழ்ந்து'],
                ['யாரிக்', 'அழ்ந்து'],
                ['யாரிகழ்ந்த்', 'உ']]
        """
        L = len(self.letters)-1
        for idx,letter in enumerate(self.letters):
            if not( letter in tamil.utf8.grantha_uyirmei_letters):
                continue
            muthal = idx == 0 and u"" or u"".join(self.letters[0:idx])
```

```
        meethi = idx == L and u"" or u"".join(self.letters[idx+1:])
        mei,uyir = tamil.utf8.splitMeiUyir(letter)
        muthal = muthal + mei
        meethi = uyir + meethi
        self.results.append([muthal,meethi])
    return len(self.results) > 0

def filter(self,lexicon):
    self.results = list( filter(lambda x: all( map(lexicon.isWord,x) ),self.results) )
    return self.results
```

## 5. Typographical error correction in Tamil

அதாவது தமிழில் தமிழ்99 அல்லது அஞ்சல் போன்ற விசைகளின் வழி உள்ளீடு செய்கையில் எழுத்துப்பிழைகள் வந்துவிடும். இதனை எப்படி சரிபாப்பது ? சில செயல்முறைகளை இதற்காக கையாளலாம். இதன் விரிவான கட்டுரையை இங்கு காணலாம் [10].

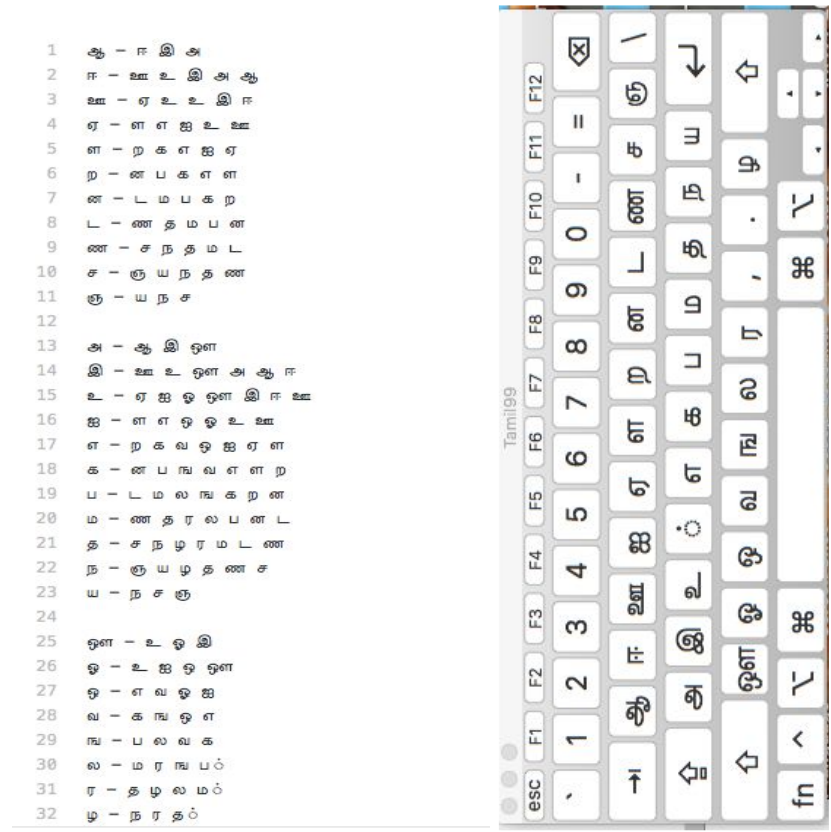

Figure: (left) confusion matrix for Tamil-99 keyboard, (right) Keyboard layout - Tamil-99.

இந்த அல்கோரிதத்தின் நிரலாக்கம் இங்கு ஓப்பன் தமிழ் திரட்டில் சேர்க்கப்பட்டது. இதனை நீங்கள் முழுதேடலில் இடம் கொடுத்தால் 2398 விடைகள் கிடைக்கும் – அதாவது முழு 4-எழுத்து சொல்லின் 4-எழுத்து தொலைவில் உள்ள திருத்தங்கள் எல்லாவற்றையும் தேடுவதால் உண்டாகும் தகவல் வெள்ளப்பெருக்கு; சாதாரணமாக 1 அல்லது 2 எழுத்துப்பிழைகள் மட்டுமே உள்ளன என்பது அறிவியலாளர்கள் கணிப்பு. இதை நாம் செயல்படுத்தும் 'tree pruning search' அல்கோரிதம்

வகையினால் நாம் 56 மாற்றங்களுக்குள் மட்டுமே தேடல்களை நடத்தி இந்த தட்டச்சு கைவிரல் தவரான உள்ளூட்டிற்கு தீர்வு காணலாம்.

இதன் சிக்கல் அளவு [computational complexity] என்பது, ஒரு n-எழுத்து சொல் என்று கொண்டால், $O(k_1 \times k_2 \times k_3 \ldots k_n) = O(k^n)$ என்று அதிக பட்சமாக இருக்கலாம் என்று [ஏதோ ஒரு k > 0 எண்ணால்] என்று நம்மால் காட்டமுடியும். This is also a variant of edit-distance search.

```python
# explore all edit distances - i.e. len(word_in) or only upto value in ed.
# we can restrict the edit distance search to any value from [1-N]
def oridam_generate_patterns(word_in,cm,ed=1,level=0,pos=0,candidates=None):
    """ ed = 1 by default, pos - internal variable for algorithm'' """
    alternates = cm.get(word_in[pos],[])
    if not candidates:
        candidates = []
    assert ed <= len(word_in), 'edit distance has to be comparable to word size [ins/del not explored]'
    if (pos >len(word_in)) or ed == 0:
        return candidates
    pfx = ''
    sfx = ''
    curr_candidates = []
    for p in range(0,pos):
        pfx = pfx + word_in[p]
    for p in range(pos+1,len(word_in)):
        sfx = sfx + word_in[p]
    for alt in alternates:
        word_alt = pfx + alt + sfx
        if not (word_alt in candidates):
            candidates.append( word_alt )
            curr_candidates.append( word_alt )
    for n_pos in range(pos,len(word_in)):
        # already what we have ' candidates ' of this round are edit-distance 1
        for word in curr_candidates:
            oridam_generate_patterns(word,cm,ed-1,level+1,n_pos,candidates)
    if level == 0:
        #candidates.append(word_in)
        for n_pos in range(pos,len(word_in)):
            oridam_generate_patterns(word_in,cm,ed, level+1,n_pos,candidates)
    return candidates

def corrections(word_in,dictionary,keyboard_cm,ed=2):
    """
    @input: word_in - input word
        dictionary - dictionary/lexicon
        keyboard_cm - confusion matrix for keyboard in question
    """
    assert isinstance(dictionary,Dictionary)
```

```python
    candidates = oridam_generate_patterns(word_in,keyboard_cm,ed)
    #TBD: score candidates by n-gram probability of language model occurrence
    #etc. or edit distance from source word etc.
    return list(filter(dictionary.isWord,candidates))
```

## 6. AI/ML approaches to Tamil spelling correction

Wide-spread success of various tasks like image recognition [6] which surpasses human level cognition in this task, and significant improvements in speech synthesis and speech recognition have demonstrated A.I. and Machine Learning methods suitable for these tasks, Sequence to sequence models including Recurrent Neural Networks (RNN), Long Short-term Memory (LSTM), and Word2Vec type representations of Neural Networks have enabled high success rate in translation, concordance tasks [7].

## 6.1 Anomaly Detection to Reduce Processing

Currently spelling correction is required in less than 25% of user input which remains perhaps requiring deep analysis and complex algorithms. To improve speed of spelling correction we can also take the approach of training a AI/ML based system to declarate word or text as valid Tamil word - this can (in principle) automatically resolve 75% of input text and free up the time required for the complex analysis. Such heuristics can allow using anomaly detection algorithms for proper identification of the errors in the text and filter the computational load on the spell checking program.

Further, Tamil misspelled letters can be collected using crowd-sourced format and used as training data for a Tamil based Deep Neural Network (DNN) for the purpose of classifying correctly spelled and misspelled words. Further, Tamil verb declension problem (see: Rajam Krishnan) can also perhaps be resolved using the big-data AI/ML approaches.

## 6.2 Word Sense Disambiguation

1) அன்பே சிவம். 2) அன்பே சவம்

Both the sentences contain legal words from a Tamil Lexicon but only one makes sense [12]. The simple minded non-word error correcting spell-checker will not be able to tell them apart.The simple way to tell them apart is to use a concordance database for Tamil and find the words co-occurring successors/predecessors and offer alternate. Yet another way is to find the word-level bi-gram, tri-gram probability distribution from a language model for Tamil and use it to identify சவம் is not a highly probable successor to 'அன்பே' thereby determining such an instance could trigger the suitable search and replacement.

## 6.3 Algorithm for Foreign-Language Word Substitution

In a previous research article we proposed a fully feed-forward ANN which was capable of identifying non-tamil words in text as well as English text (originally reported in [17]).

We propose a simple algorithm which can use the above AI classifier and extract this word and look up equivalent Tamil word for English or other language text using a parallel dictionary. Updating the word for the tense and any morphological processing we can replace the anglicized word in Tamil script with an equivalent Tamil word.

## 6.4 Computational Complexity

We try to answer the question of complexity of Tamil language spell-checker algorithm in various contexts. It seems the correction of Tamil words in text will be somewhere between exponential in number of letters of word - clearly the substitution cases are shown to be exponentially complex in number of letters.The complexity of affix removal is polynomial in the number of affixes.

# 7 Optimizing a Spell Checker Implementation

We can propose techniques for how to improve speed of spell checker.

## 7.1 Algorithm for fast Unicode letter detection

Text containing a mix of Tamil and English (or other language scripts) can be quickly eliminated by using the check if a character falls into the Tamil Unicode block in basic block range. This check eliminates a character in analyzed text from further complex processing. In Python3, the following function performs the Unicode Tamil letter/character detection.

```
is_tamil_unicode_predicate = lambda x: x >= chr(2946) and x <= chr(3066)
```

## 7.2 Performance Engineering

### 7.2.1 Caching results

A given text for spelling correction can be grouped into a series of words, excluding stop words. Now we can make spell checker save suggestion list for each mis-spelled word (non-word) and re-use the suggestions list from cache the second and later times when word is mis-spelled identically in the document. The same mis-spelling can, however, have different correct words in the document at each site depending on context and only be replaced upto first order approximation in "white-washed" fashion. Caching has a potential to restore performance of spell checker.

### 7.2.2 Multi-Threading/Multi-Processing

The non-word errors suggestion generation and morphological processing can be carried out in parallel in the whole document, perhaps by a producer-consumer model serviced by a dozen(s) of worker threads which all generate suggestions for non-word errors including caching as mentioned above. Such an architecture of spelling checker could improve speed.

## 7.2.3 Redis / distributed DB

The memory performance of the spelling checker can be offloaded to a remote machine by requesting a distributed data-based to hold the dictionary or trie form of the word-list/Lexicon. This may be suitable for production environments.

# 8 Conclusion

In this paper we have attempted to make a summary of various known algorithms for specific classes of Tamil spelling errors. We believe this collection of suggestions to improve future spelling checkers. We also note do not cover many important techniques like affix removal and other such techniques of key importance in rule-based spell checkers.